\documentclass[runningheads]{llncs}
\usepackage{graphicx}
\usepackage{caption}
\usepackage{subcaption}
\usepackage{stfloats}

\usepackage{xspace}  % Used only for xspace
\usepackage{booktabs}  % Top rule, bottom rule...
\usepackage{xcolor}  % To color text.
\usepackage{hyperref}

\newcommand{\TaskA}{Single Document VQA\xspace}
\newcommand{\TaskB}{Document Collection VQA\xspace}
\newcommand{\TaskC}{Infographics VQA\xspace}

\newcommand*\samethanks[1][\value{footnote}]{\footnotemark[#1]}

\hypersetup{
    colorlinks=true,
    linkcolor=blue,
    filecolor=magenta,      
    urlcolor=blue,
}
\begin{document}
\title{ICDAR 2021 Competition on Document Visual Question Answering}
% \title{ICDAR 2021 Long-term Competition on DocVQA}
\author{Rub{\`e}n Tito\thanks{Equal contribution.}\inst{,1}
\and Minesh Mathew\samethanks{}\inst{,2} 
\and C.V. Jawahar\inst{2} \and Ernest Valveny\inst{1} \and Dimosthenis Karatzas\inst{1}}
% \and C.V. Jawahar\inst{2} \and Ernest Valveny\inst{1} \and R. Manmatha\inst{3} \and Dimosthenis Karatzas\inst{1}}

\authorrunning{R. Tito et al.}

\institute{Computer Vision Center, UAB, Spain \\
\email{\{rperez, ernest, dimos\}@cvc.uab.cat} \and
CVIT, IIIT Hyderabad, India
%\email{minesh.mathew@research.iiit.ac.in}
}

\maketitle              % typeset the header of the contribution
%

%% commands for commonly used names %%
\newcommand{\datasetname}{InfographicsVQA\xspace}
\newcommand{\docvqa}{DocVQA\xspace}
\newcommand{\doccvqa}{DocCVQA\xspace}
\newcommand{\squad}{SQuAD\xspace}
\newcommand{\vqa}{VQA 2.0\xspace}
\newcommand{\textvqa}{TextVQA\xspace}
\newcommand{\stvqa}{ST-VQA\xspace}
\newcommand{\ocrvqa}{OCR-VQA\xspace}

\newcommand{\lorra}{LoRRA\xspace}
\newcommand{\mc}{M4C\xspace}
\newcommand{\bertbasemodel}{\texttt{bert-base-uncased}\xspace}
\newcommand{\bertlargemodel}{\texttt{bert-large-uncased-whole-word-masking}\xspace}
\newcommand{\bertlargesquadmodel}{\texttt{bert-large-uncased-whole-word-masking-finetuned-squad}\xspace}
\newcommand{\layoutlm}{LayoutLM\xspace}
\newcommand{\layoutlmtwo}{LayoutLM 2.0\xspace}
%TAP

\newcommand{\visualmrc}{VisualMRC\xspace}
\newcommand{\drop}{DROP\xspace}
\newcommand{\dvqa}{DVQA\xspace}
\newcommand{\leafqa}{LEAF-QA\xspace}
\newcommand{\figureqa}{FigureQA\xspace}
\newcommand{\textbookqa}{TQA\xspace}
\newcommand{\recipeqa}{RecipeQA\xspace}

\newcommand{\visuallydataset}{Visually29K\xspace}

% Task 2 methods.
\newcommand{\infrarrd}{Infrrd-RADAR\xspace}

% Task 3 methods.
\newcommand{\tilt}{Applica.ai TILT\xspace}
\newcommand{\igbert}{IG-BERT (single model)\xspace}
\newcommand{\clova}{NAVER CLOVA\xspace}
\newcommand{\ensemble}{Ensemble LM and VLM\xspace}
\newcommand{\bertba}{bert baseline\xspace}
\newcommand{\bert}{BERT (CPDP)\xspace}  % Commands for commonly used names %%

\begin{abstract}
In this report we present results of the ICDAR 2021 edition of the Document Visual Question Challenges. This edition complements the previous tasks on \TaskA and \TaskB with a newly introduced on \TaskC.
% In this report we present results of the ICDAR 2021 edition of the Document Visual Question Challenges. This edition of the challenge featured two tasks --- a VQA task where questions are asked over a document collection of documents having same template (\TaskB) and a newly introduced task on VQA on infographics (\TaskC). 
\TaskC is based on a new dataset of more than $5,000$ infographics images and $30,000$ question-answer pairs. % \TaskB is adapted from the 2020 challenge to rank the submitted methods according to their answering performance instead of the retrieval one, highlighting the VQA aspect of this task and leaving the retrieval of the positive evidences as a robust explanation of where the answers have been inferred from.
% Methods tackling Visual Question Answering on documents has evolved following the trend of using pretrained language representations with multi-modal transformers since most of the top ranked methods in \TaskA and \TaskC follow this approach. % The winner method of \TaskC task, which is also the currently best ranked method in \TaskA task is based on multi-modal transformers and relies on an encoder-decoder architecture to generate words that are not included in the input text. On the contrary, the winner of \TaskB task extracts the key information of the whole document collection to then retrieve the relevant documents and extract the answer.
The winner methods have scored $0.6120$ ANLS in \TaskC task, $0.7743$ ANLSL in \TaskB task and $0.8705$ ANLS in \TaskA. We present a summary of the datasets used for each task, description of each of the submitted methods and the results and analysis of their performance.
A summary of the progress made on \TaskA since the first edition of the DocVQA 2020 challenge is also presented.

\end{abstract}
\section{Introduction}
% Machine perception of images is an important goal towards building intelligent AI systems. Different from traditional Computer Vision problems focused on detection and recognition of images, objects or scenes, this requires high level semantic understanding of images.
% Computer Vision community has actively been pursuing this direction by building models with Visual Question Answering (VQA) ability~\cite{vqa1,vqa2,johnson2017clevr,topdown_bottomup}.  
Visual Question Answering (VQA) seeks to answer natural language questions asked on images. The initial works on VQA focused primarily on images in the wild or natural images~\cite{vqa1,gqa}. Most models developed to perform VQA on natural images, make use of (i) deep features from whole images or objects (regions of interest within the image), (ii) a Question Embedding module which make use of word embeddings and Recurrent Neural Networks (RNN) or Transformers~\cite{attention_is_all_you_need}, and (iii) a fusion module which fuses the image and text modalities using attention~\cite{topdown_bottomup,vqa_tips}.

Images in the wild often contain text and reading this text is critical to semantic understanding of natural images. For example, Veit et al. observe that nearly 50\% of images in MS-COCO dataset have text present in it~\cite{cocotext}. TextVQA~\cite{singh2019towards} and ST-VQA~\cite{biten2019scene} datasets evolved out of this actuality and both the datasets feature questions where reading text present on the images is important to arrive at the answer. Another track of VQA which require reading text on the images is VQA on charts. DVQA~\cite{dvqa}, FigureQA~\cite{figureqa} and LEAF-QA~\cite{leafqa} comprise a few types of standard charts such as bar charts or line plots. Images in these datasets are rendered using chart plotting libraries using either dummy data or real data and the datasets contain millions of questions created using question templates.

On the other hand, Reading Comprehension~\cite{rajpurkar2016squad,triviaqa,drop} and Open-domain Question Answering~\cite{ms_marco,natural_quesions} on machine readable text is a popular research topic in Natural Language Processing (NLP) and Information Retrieval (IR). Unlike VQA where the context on which the question is asked is an image, here questions are asked on a given text passage or a text corpus, say the entire Wikipedia. Recent developments in large-scale pretraining of 
language models using huge corpora of unlabeled text and later transferring pretrained representations to downstream tasks enabled considerable progress in this space~\cite{bert,xlnet}.

Similar to QA involving either images or text, there are multimodal QA tasks where questions are asked on a context of both images and text. Textbook QA~\cite{textbookqa} is a QA task where questions are based on a context of text and diagrams and images. RecipeQA~\cite{recipeqa} questions require models to read text associated with a recipe and understand related culinary pictures. Both datasets have multiple choice style questions and the text modality is presented as machine readable text passages, unlike unstructured text in VQA involving scene text or chart VQA, which need to be first recognized from the images using an OCR.

% DK TRIAL
Document analysis and recognition aims to automatically extract information from document images. DAR research tends to be bottom up and focus on generic information extraction tasks (character recognition, layout analysis, table extraction, etc), disconnected from the final purpose the extracted information is used for.
The DocVQA series of challenges aims to bridge this gap, by introducing Document Visual Question Answering (DocVQA) as a high-level semantic task dynamically driving DAR algorithms to conditionally interpret document images.
Automatic document understanding is a complicated endeavour that implies much more than just reading the text. In designing a document the authors order information in specific layouts so that certain aspects “stand out”, organise numerical information into tables or diagrams, request information by designing forms, and validate information by adding signatures. All these aspects and more would need to be properly understood in order to devise a generic document VQA system.
% END DK TRIAL

% Document images while stored and transmitted as images, carry a lot of textual information. At the same time documents can have complex layouts and graphics which make the Computer Vision aspect challenging. For example, a document image could be a scanned image of a single column text from a book or a scanned copy of a newspaper or magazine with a complex layout and rich interplay of text, images, tables and visualizations.
% Inspired by the recent interest on VQA on Images containing embedded text and multimodal QA, we introduced the Document Visual Question Answering (DocVQA) challenge as part of the `Text and Documents in the Deep Learning Era' workshop in CVPR 2020, from which a short report was published in~\cite{docvqa_das}. We envision to organize a series of challenges, workshops and build datasets and models for Question answering on document images. More details about the datasets, the future challenges and updates can be found in \href{https://docvqa.org}{https://docvqa.org}.
 
% \input{figures/three_tasks_samples}
\begin{figure*}[ht]
    \renewcommand{\arraystretch}{2} % Default value: 1
    \centering
    \begin{tabular}{ccc}
    \begin{subfigure}{0.33\textwidth}
        \includegraphics[width=\textwidth]{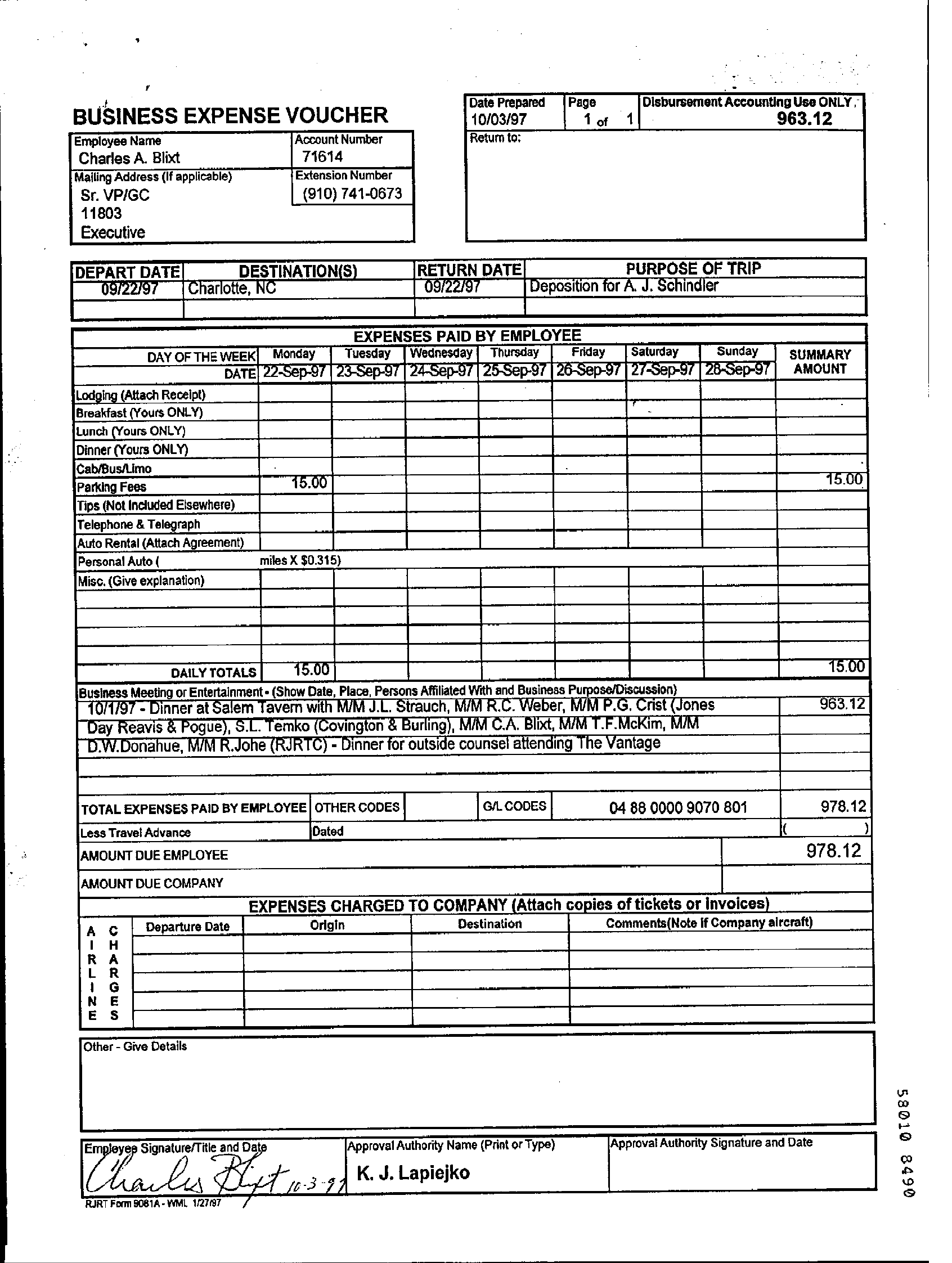}
        % {\footnotesize \textbf{Question:} What is the extension number as per the voucher? \par
        %     \textbf{Answers:} (910) 741-0673}
      \end{subfigure} &
      
     \begin{subfigure}{0.33\textwidth}
        \includegraphics[width=\textwidth]{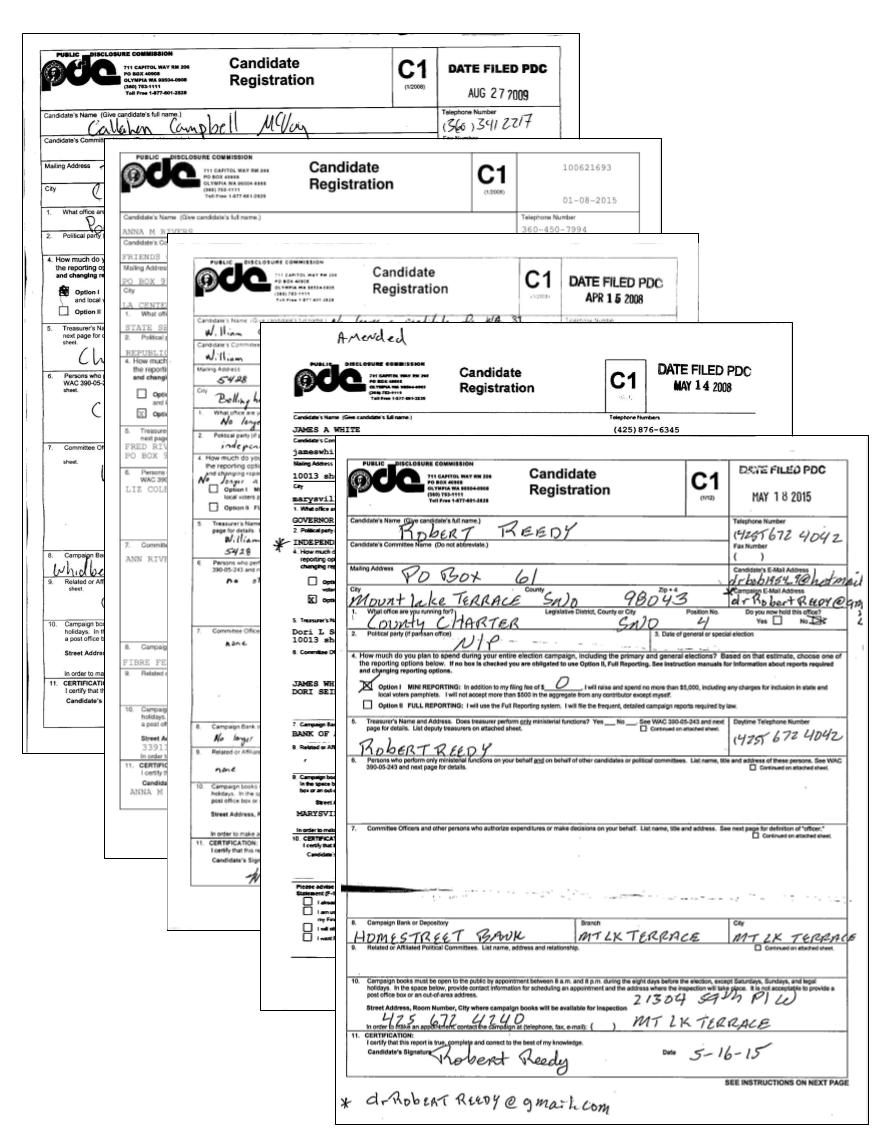}
      \end{subfigure} &
      
     \begin{subfigure}{0.33\textwidth}
        \includegraphics[width=\textwidth]{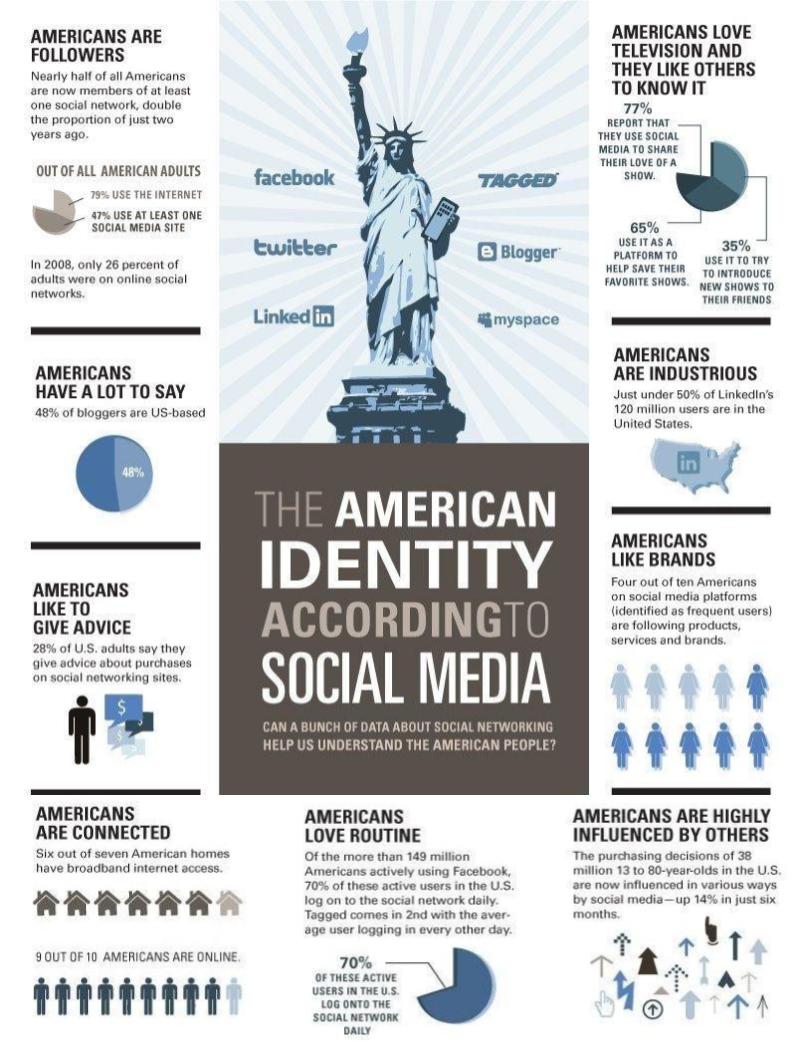}
      \end{subfigure}  \\
      
    \begin{minipage}{0.30\textwidth}
            \vspace{10pt}
            \scriptsize \textbf{Question:} What is the extension number as per the voucher? \par
            \textbf{Answers:} (910) 741-0673
    \end{minipage} & 
    
    \begin{minipage}{0.30\textwidth}
            \vspace{15pt}
            \scriptsize \textbf{Question:} In which years did Anna M. Rivers run for the State senator office? \par
            \textbf{Answers:} 2016, 2020 \par
            \textbf{Doc. Evidences:} 454, 10901
    \end{minipage} & 
    
    \begin{minipage}{0.30\textwidth}
            \scriptsize \textbf{Question}: What percentage of Americans are online? \par 
            \textbf{Answers:} 90\%, 90
    \end{minipage} \\
    
    \begin{subfigure}{0.33\textwidth} \caption{\scriptsize\TaskA} \label{subfig:Task_ex_T1} \end{subfigure} &
    \begin{subfigure}{0.33\textwidth} \caption{\scriptsize\TaskB} \label{subfig:Task_ex_T2} \end{subfigure} &
    \begin{subfigure}{0.33\textwidth} \caption{\scriptsize\TaskC} \label{subfig:Task_ex_T3} \end{subfigure} 
    
    \end{tabular}
    \caption{Examples for the three different tasks in DocVQA. \TaskA task (a) is a standard VQA on business/industry documents. Instead, in \TaskB task (b), questions are asked on a document collection comprising documents of same template. Finally, \TaskC task (c) is similar to \TaskA task, but the images in the dataset are infographics.}
    \label{fig:three_taks_examples}
\end{figure*}

The DocVQA challenge comprises three tasks. The \TaskA and \TaskB tasks were introduced in the 2020 edition of the challenge as part of the `Text and Documents in the Deep Learning Era' workshop in CVPR 2020~\cite{docvqa_das}, and have received continuous interest since then. We discuss here important changes and advancements in these tasks since the 2020 edition. The \TaskC task is a newly introduced task for the 2021 edition.

The \TaskA task %, originally introduced in the 2020 edition of DocVQA, 
requires answering questions asked on a single-page document image. The dataset for \TaskA comprises business documents, which include letters, fax communications reports with tables and charts and others. These documents mostly contain text in the form of sentences and passages~\cite{docvqa_wacv}. %The \TaskA task was introduced for the 2020 edition of DocVQA and has received continuous interest since then. 
We discuss here new submissions for this task since the 2020 edition.

In the \TaskB task, questions are posed over a whole collection of documents which share the same template but different content. More than responding with the right answer, in this scenario it is important to also provide the user with the right evidence in collection to support the response.
For the 2021 edition, we have revisited the evaluation protocols to better deal with unordered list answers and to provide more insight into the submitted methods.
%In that edition the ranking of the methods was according to the retrieval of the relevant documents to answer the question letting the answer itself as an optional response. On this ICDAR 2021 version we have updated the task to rank the methods according to the answering performance and leaving the retrieval of the positive evidences as an answer robustness indicator. 

%Finally, the \TaskC task is a newly introduced task in this edition of the DocVQA challenge. Questions 
In the newly introduced \TaskC task, questions are asked on a single image, but unlike \TaskA  where textual content is dominating the document images, here we focus on infographics, where the structure, graphical and numerical information are also important to convey the message.
%where images are business documents, images here are infographics. 
For this task we introduced a new dataset of infographics and associated questions and answers annotations~\cite{infographicvqa}.

% We provide a service that encompasses all DocVQA framework from all the tasks that allows to explore the datasets as well as some of the implemented methods in \href{https://docvqa.org}{https://docvqa.org}. On the other hand, we provide all the challenge related information in \href{https://rrc.cvc.uab.es/?ch=17}{https://rrc.cvc.uab.es/?ch=17}.

% The rest of the paper is organized as follows. Section \ref{comp.protocol} describes the competition protocol, such as the competition platform challenge date and other details. In sections following it we discuss each task in detail, one task per section, starting with the new task -- \TaskC.

 More details about the datasets, the future challenges and updates can be found in \href{https://docvqa.org}{https://docvqa.org}.

\section{Competition Protocol}
\label{comp.protocol}

The Challenge ran from November 2020 to April 2021. 
%In Task 1, as a long-term continuation of the competition nothing in the challenge setup was modified. In the Task 2 the dataset was the same as the previous edition, but the evaluation metric was changed from MAP to ANLSL (section \ref{subsec: task2_metric}) to rank the methods according to the answering performance instead of the retrieval of the evidences. 
The setup of the \TaskA and \TaskB tasks was not modified with respect to the 2020 edition, while for \TaskC, which is a completely new task, we released the training and validation sets between November 2020 and January 2021, and the test set on February 11, 2021, giving participants two months to submit results until April 10th.
%a first subset with half of the training set was released on November 10th and the other half on December 23rd. The validation set was released on January 5th and the test set on February 11th. The submission deadline for all the tasks was set on April 10th. The participants were requested to submit their results on the test split of the dataset.
We relied at all times on the scientific integrity of the authors to follow the established rules of the challenge that they had to agree with upon registering at the Robust Reading Competition portal.
%Along with each dataset split released we also provided the image OCR tokens extracted by Amazon Textract. 

The Challenge is hosted at the Robust Reading Competition (RRC) portal\footnote{\href{https://rrc.cvc.uab.es/?ch=17}{https://rrc.cvc.uab.es/?ch=17}}.
%which was developed in 2011 to host the original robust reading competitions concerning text detection and recognition from born-digital and scene images. Since then it has evolved to a fully-fledged platform for hosting academic contests. At the time of running this challenge, the portal hosts $17$ different challenges, structured in $49$ different tasks. The platform currently has more than $24,000$ registered users from over $139$ countries, with more than $64,800$ methods evaluated to date. 
All submitted results are evaluated automatically, and per-task ranking tables and visualization options to explore results are offered through the portal.
The results presented in this report reflect the state of submissions at the closure of the 2021 edition of the challenge, but the challenge will remain open for new, out-of-competition, submissions. The RRC portal should be considered as the archival version of results, where any new results, submitted after this report was compiled will also appear.

\section{\TaskC}
\label{sec:Task3}
% Based on our learnings from Task1 , we designed a new DoCVqA task 3, which is similar to Task1 but this time on infographics. We wanted this VQA task to have questions which makes the models understand the layout and visual aspect of the image. We argue that infographics which are rich in graphics will be a perfect testbed for VQA capability involving text, vision and layout. This edition of the challenge included a competition on the new Task3 and continued Task2 as well.
The design of the new task on \TaskC was informed by our analysis of results from the 2020 edition of the \TaskA task. In particular, we noticed that most state of the art methods for \TaskA followed a simple pipeline consisting on applying OCR (that was also provided) followed by purely text-based QA approaches. Although such approaches give good overall results as running text dominates the kind of documents of the \TaskA task, a closer look at the different question categories reveals that those methods performed quite worse on questions that rely on interpreting graphical information, handwritten text or layout information. In addition, these methods work because the \TaskA questions were designed so that they can be answered in an extractive manner, which means that the answer text  appears in the document image and usually can be inferred from the surrounding  context.

For the new \TaskC task, we focused on a domain where running text is not dominating the document, while we downplayed the amount of questions based on purely textual evidence and defined more questions which require the models to interpret the layout and other visual aspects. We also made sure that a good number of questions require logical reasoning or elementary arithmetical operations to arrive at the final answer.

We have received and evaluated a total of $337$ different submissions on this task from $18$ different users. Those submissions include different versions from the same methods. % and private and test submission, better not to say... no?
In section \ref{subsec:task3_submitted_methods} can be found the ranking of the $6$ methods that finally participated in the competition.
% Unlike in task 1, where answers to all questions are a contiguous piece of text from the document image ( a `span') we allow for non-span and multi-span answers in task 3. Specifically answer to a question can be (i) a `span' of text from the infographic image in the reader (ii) multiple spans from the image or (ii) a numerical answer which is not extracted from the image.

\subsection{Evaluation Metric}
To evaluate and rank the submitted methods on this task we use the standard ANLS~\cite{biten2019icdar} metric for reading-based VQA tasks. There exist a particular case in \TaskC task where the full answer is actually a set of items for which the order is not important, like a list of ingredients in a recipe. In the dataset description (section \ref{subsec: infovqa dataset}) we name this as 'Multi-Span' answer.
In this case we create all possible permutations of the different items and accept it as correct answers.

\subsection{The \datasetname Dataset}
\label{subsec: infovqa dataset}

Infographic images were downloaded from various online sources using the Google and Bing image search engines. We removed duplicate images using a Perceptual Hashing technique implemented in Imagededup library~\cite{imagededup}.
% To gather the question-answer pairs we used an internal web based annotation tool.
% Instead of using online crowdsourcing platforms, we contracted annotators and worked closely with them through continuous interaction to ensure the annotations quality and to balance of question types defined.
To gather the question-answer pairs, instead of using online crowdsourcing platforms we used an internal web based annotation tool, hired annotators and worked closely with them through continuous interaction to ensure the annotations quality and balance the amount of questions for each type defined.
In cases where there are multiple correct answers due to language variability, we collected more than one answer per question. For example in the case of the example shown in Figure~\ref{subfig:Task_ex_T3} the question has two valid ground-truth answers.

% The annotation process was split in two stages. In the first one, employed annotators were asked to define questions and answers over the images. 
% We allowed them to add multiple answers per question to better capture the variability while answering. For example, if the question asks about a percentage, both answers $50$, or $50\%$ shall be considered correct. \redText{Check if we can use some image example to make this example more clear.} 
% The second stage of annotation was essentially a validation process for images in the validation and test sets to ensure a good quality during the evaluation. In this stage the images and already defined questions were shown to the annotators (different from the stage 1).And then, they had to come up with the answer. Afterwards, the answers from the annotation and verification stages were compared. For each question, if at least one answer from each stage matched, that question and all their answers were included in the dataset. In the other case, the question and answers were rejected. 
In total $30,035$ question-answer pairs were annotated on $5,485$ infographics.
We split the data into train, validation and test splits in an 80-10-10 ratio, with no overlap of images between different splits. 
We also collected additional information regarding questions and answers in the validation and test splits which help us better analyze VQA performance. 
In particular, we collected the type of answer, which indicates whether the answer can be found as (i) a single text-span in the image (`image-span'), (ii) a concatenation of different text-spans in the image (`multi-span'), (iii) a span of the question text (`question span') or (iv) a `non-span' answer. The evidence type for each answer is also annotated, and there could be multiple evidences for some answers. The different evidence types are Text, Table/list, Figure, Map (a geographical map) and Visual/layout. Additionally, for questions where a counting, arithmetic or sorting operation is required we collect the operation type as well. More details of the annotation process, statistics of the dataset, analysis of images, questions and answers and examples for each type of evidence, answer, and operation are presented in~\cite{infographicvqa}. 

Along with the \datasetname dataset, we also provided OCR transcriptions of each of the images in the dataset obtained using Amazon Textract OCR.

\subsection{Baselines}
\label{subsec:task3_baselines}

%For this task we show results using two baselines. 
In order to establish a baseline performance, we used two models built on state of the art methods.
The first model is based on a layout aware language modelling approach --- \layoutlm~\cite{xu2020layoutlm}. We take a pretrained \layoutlm model and continue pretraining it on the train split of the \datasetname dataset using a Masked Language Modelling task. Later we finetune this model for an extractive question answering using a span prediction head at the output. We name this baseline as ``(Baseline) \layoutlm''.

The second baseline is the state of the art Scene Text VQA method %model. involving images with embedded scene text. 
\mc~\cite{hu2020iterative}. This method embeds all different modalities -- question, image text and image, into a common space, where a stack of transformer layers is applied allowing each entity to attend to inter- and intra- modality features providing them with context. Then, the answer is produced by an iterative decoder with a dynamic pointer network that can provide an answer either from a fixed vocabulary or from the OCR tokens spotted on the image. This baseline is named as ``(Baseline) \mc''.
In \cite{infographicvqa} we give more details of these models, various ablations we try out and detailed results and analysis.

We also provide a human performance evaluation carried out by two volunteers ``(Baseline) Human''.

\subsection{Submitted methods}
\label{subsec:task3_submitted_methods}
We received 6 submissions in total. All of them are based on pretrained language representations. However, we can appreciate that the top ranked methods use multi-modal pretrained architectures combining visual and textual information extracted from the image, while the lower ranked methods use representations based only on natural language.

\begin{itemize}
\setlength\itemsep{1em}

\item[\#1 -] \tilt~\cite{powalski2021going}: The winning method learns simultaneously layout information, visual features and textual semantics with a multi-modal transformer based method, and rely on an encoder-decoder architecture that can generate values that are not included in the input text.

\item[\#2 -] \igbert: A visual and language pretrained model on infographic text pairs. The model was initialized from BERT-large and trained on \datasetname training and validation data. The visual features are extracted using a Faster-RCNN trained on Visually29K~\cite{visually29k}. Also, they used OCR tokens extracted by the Google Vision API instead of the ones provided in the competition.

\item[\#3 -] \clova: This method uses and extractive QA method based on the HyperDQA~\cite{docvqa_das} approach. First they pre-train BROS~\cite{hong2021bros} model on the IIT-CDIP~\cite{iitcdip_dataset} dataset but sharing the parameters between projection matrices during self-attention. Then, they perform additional pretraining on \squad~\cite{rajpurkar2016squad} and WikitableQA~\cite{pasupat2015compositional} datasets. Finally, they also fine-tune on the \docvqa~\cite{docvqa_das} dataset.

\item[\#4 -] \ensemble: An ensemble between two different methods from extractive NLP QA method (Method 1) and scene-text VQA method (Method 2). The first one is based on BERT-large pretrained on \squad + \docvqa and fine-tuned on \datasetname. The final output is vote-based by three trained models with different hyper-parameters. The second model is the SS-Baseline~\cite{zhu2020simple} trained on TextVQA~\cite{singh2019towards}, ST-VQA~\cite{biten2019scene} and \datasetname. At the end a rule-based post-processing is performed for three types of questions, i.e., selection, inverse percentage, and summation to get the final result by filling the empty results of Model 1 with answers predicted by Model 2.

\item[\#5 -] \bertba: BERT-large model pretrained on \squad~\cite{rajpurkar2016squad}. It also uses a fuzzy search algorithm to better find the start and end index of the answer span.

\item[\#6 -] \bert: BERT-large model pretrained on \squad~\cite{rajpurkar2016squad}.
\end{itemize}

\vspace{-20pt}

\setlength{\tabcolsep}{8pt}  % Increase the separation space between columns.

\begin{table}[h]
\centering
\begin{tabular}{lr}
\toprule
Method          & \multicolumn{1}{c}{ANLS} \\
\midrule
(Baseline) Human            & 0.9800             \\
\tilt                       & \textbf{0.6120}   \\
\igbert                     & 0.3854            \\
\clova                      & 0.3219            \\
\ensemble                   & 0.2853            \\
(Baseline) \layoutlm        & 0.2720             \\
\bertba                     & 0.2078            \\
\bert                       & 0.1678            \\
(Baseline) \mc              & 0.1470             \\
\bottomrule
\end{tabular}
\vspace{2mm}
\caption{\textbf{\TaskC task results table.} Top section show methods withing the ICDAR 2021 competition. Bottom section shows baseline methods proposed by the competition organizers.}
\label{tab:task3_results}
\end{table}

\setlength{\tabcolsep}{6pt}  % Restore column separation to its default value.

\vspace{-30pt}

\begin{figure}[hp]
    \centering
    \begin{subfigure}[t]{1\textwidth}
        \centering
        \includegraphics[width=\linewidth, height=5cm]{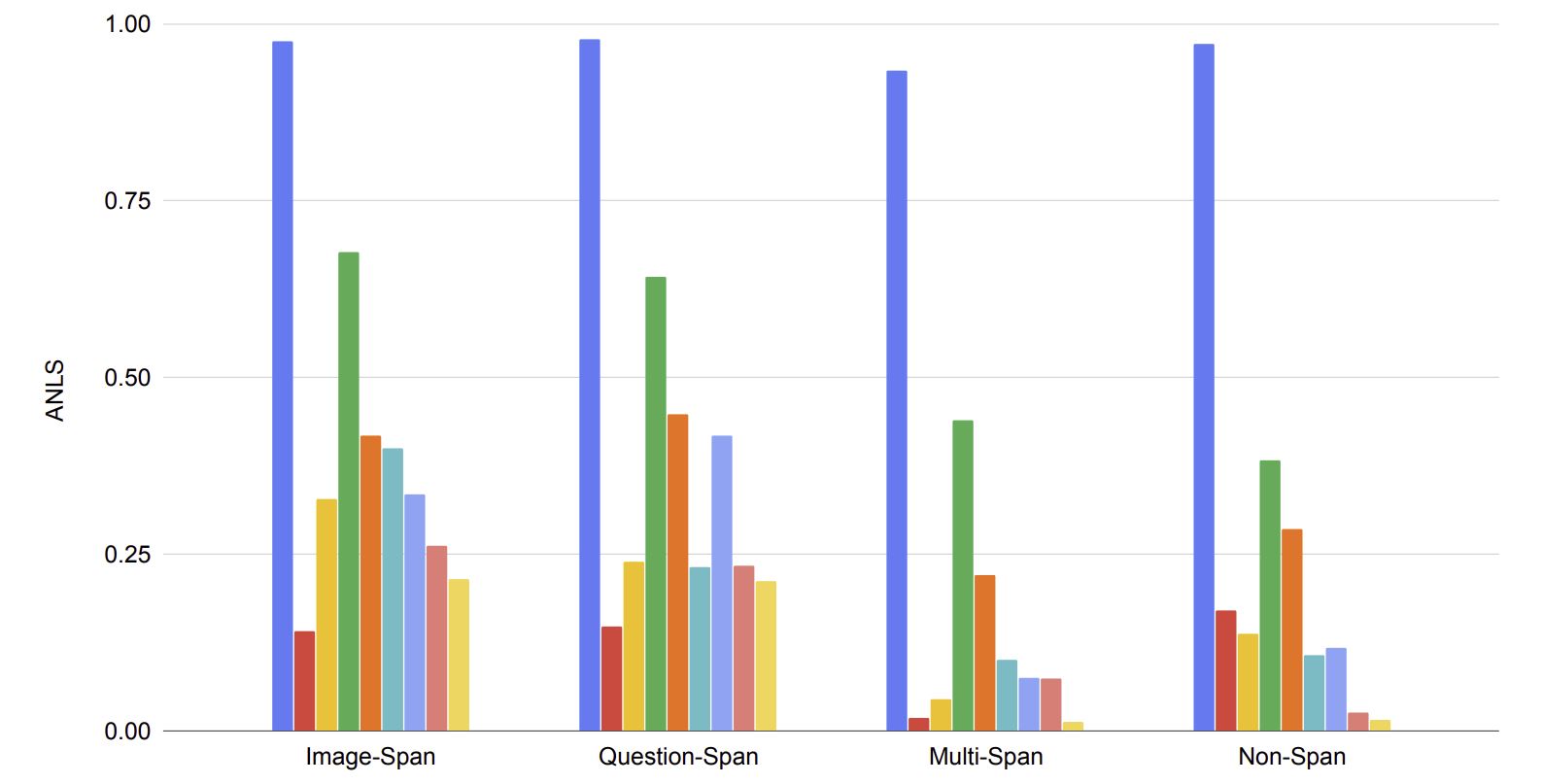}
        % \caption{question type}
        \label{fig:infovqa_question_types_results}
    \end{subfigure}
    
    \begin{subfigure}{1\textwidth}
        \centering
        \includegraphics[width=\linewidth, height=5cm]{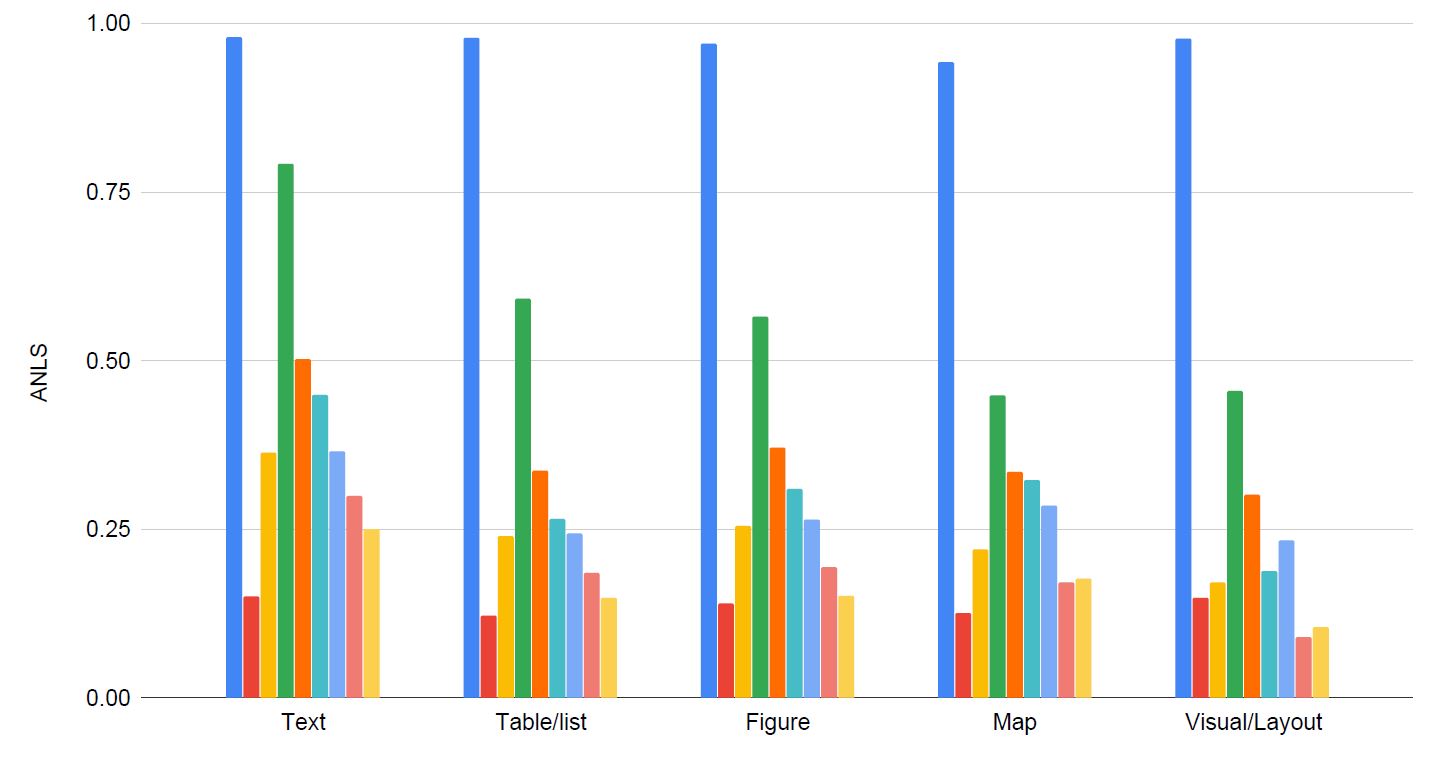}
        % \caption{evidence type}
        \label{fig:infovqa_evidence_types_results}
    \end{subfigure}
    
    \begin{subfigure}[b]{1\textwidth}
        \centering
        \includegraphics[width=\linewidth, height=5cm]{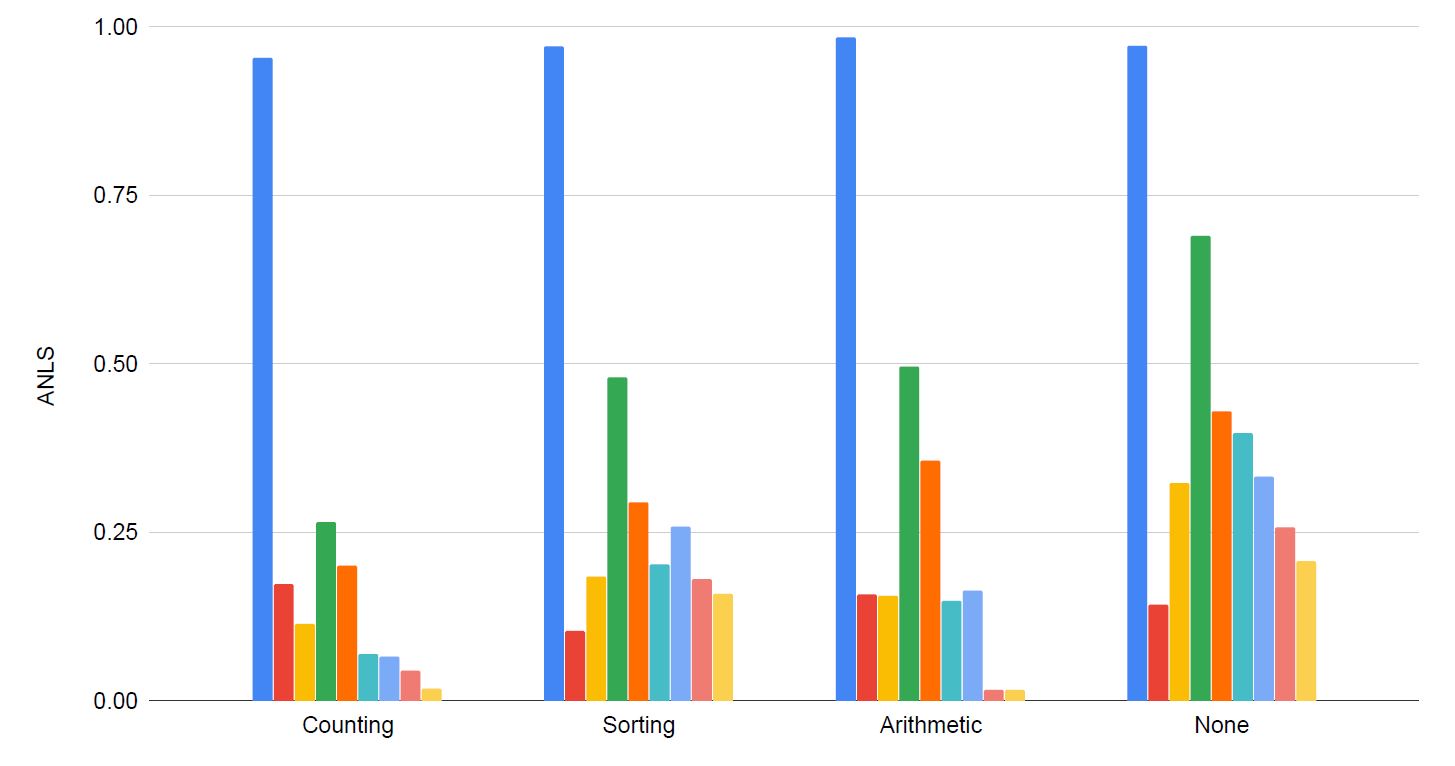}
        % \caption{operation type}
        \label{fig:infovqa_operation_types_results}
    \end{subfigure}
    \includegraphics[width=\linewidth]{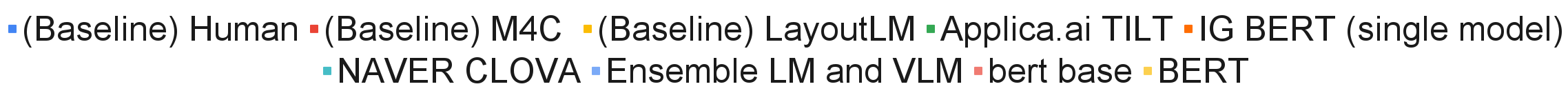}
    % \includegraphics[width=\linewidth, height=5cm]{images/plots/T3_Legend.JPG}
    % \vspace{-10.0mm}
    \caption{\textbf{ANLS score of \TaskC task methods} breakdown by answer types (top), evidence type (middle) and operations required to answer the posed questions (bottom).}
    \label{fig:infovqa_breakdown}
\end{figure}

\subsection{Performance analysis} \label{subsec_task3_results}

In table \ref{tab:task3_results} the final competition ranking of the \TaskC task is shown. In addition, in Figure~\ref{fig:infovqa_breakdown} we show the methods' Average Normalized Levenshtein Similarity (ANLS) score breakdown by type of answer (top), type of evidence (middle) and type of operations and reasoning required to answer the posed questions (bottom). As it can be seen, the winning method \tilt outperforms the rest in all the categories. In addition, the plots show a drop in performance of all methods when the text of the answer is a set of different text spans in the image (Multi-Span) or it does not appear in the image (Non-Span). The easiest questions are the ones that can be directly answered with running text from the documents (Figure~\ref{fig:infovqa_breakdown}-middle: Text), this should be expected given the extractive QA nature of most of the methods. Finally, the hardest ones are all the questions that require to perform an operation to come up with the answer as the last plot illustrates comparing the Counting, Sorting and Arithmetic against the performance on None.

\section{\TaskB}
\label{sec: task2}

In \TaskB task, the questions are posed over a whole collection of 14K document images. The objective in this task is to come up with the answers to the given questions, but also with the positive evidences. We consider positive evidence the document IDs from which the answer can be inferred from.

This task was first introduced in the scope of the CVPR 2020 DocVQA Challenge edition, and we have received and evaluated $60$ submissions from $7$ different users, out of which $5$ have been made public by their authors and feature in the ranking table. $3$ of these submissions are new, tackling the problem of answering the questions and not only providing the evidences.

\subsection{Evaluation Metric} \label{subsec: task2_metric}
% In the first edition of this task the ranking of the methods was according to the Mean Average Precision (MAP), evaluating only the retrieval of the positive evidences and relegating the answering part as an optional response. Instead, for this ICDAR 2021 edition the evaluation metric has been changed to rank the methods according to the answering performance and using the provided evidences by the methods as a robustness indicator of the answers.
In this task, the whole answer to a given question is usually a set of sub-answers extracted from different documents, and consequently the order in which those itemized sub-answers are provided is not relevant. Thus, it presents a problem similar to the Multi-Span answers in \TaskC. However, the amount of items in this task varies from $1$ to $>200$ depending on the question, and therefore is not smart to just create all possible permutations. For this reason the metric used to assess the answering performance in this task is the Average Normalized Levenshtein Similarity for Lists (ANLSL) presented in \cite{doccvqa_icdar}, an adaptation of the ANLS~\cite{biten2019icdar}, to work with a set of answers for which the order is not important. Thus, captures smoothly OCR recognition errors and evaluates reasoning capability at the same time while handles unordered set of answers by performing the Hungarian Matching algorithm between the method's provided answer and the ground truth.

In addition, even though the retrieval of the positive evidences is not used to rank the methods, we evaluate them according to the Mean Average Precision (MAP) which allows to better analyze the method's performance.

% Previous Evaluation Metric version -- The metric used to assess the answering performance is an adaptation of the ANLS~\cite{biten2019icdar}, to work with a set of answers for which the order is not important. Thus, captures smoothly OCR recognition errors and evaluates reasoning capability at the same time while handles unordered set of answers. We named this adaptation as Average Normalized Levenshtein Similarity for Lists (ANLSL).

\subsection{Baselines}

We defined two baseline methods to establish the base performance for this task.
%from different perspectives to showcase how would perform already existing methods. 
Both baselines work in two steps. First they rank the documents in the collection to find which ones are relevant to answer the given question and then, they extract the answer from those documents marked as relevant. The first method ``(Baseline) TS - BERT" ranks the documents by performing text spotting of specific words in the question over the documents in the collection. Then using an extractive QA pretrained BERT model extracts the answers from the top ranked documents. The second method named ``(Baseline) Database'', makes use of the commercial OCR Amazon Textract to extract the key-value pairs from the form's fields for all the documents in the collection. With this information a database-like structure is built. Then, the questions are manually parsed into Structured Query Language (SQL) from which the relevant documents and answers are retrieved. Note that while the first baseline is generic in nature and could potentially be applied on different collections, the second one relies on the nature of the documents (forms). Detailed information for both methods can be found in \cite{doccvqa_icdar}.

\subsection{Submitted methods}
In this task, only one new method has submitted. The winning method \infrarrd~(Retrieval of Answers by Document Analysis and Re-ranking) first applies OCR to extract textual information from all document images and the combines results with image information to extract key information. % such as the candidate name or the party. 
The question is parsed into SQL queries by using the spaCy library to split and categorize the question chunks into predefined categories. The SQL queries are then used to retrieve a set of relevant documents and BERT-Large is used to re-rank them again. Afterwards, the re-ranked document IDs are used to filter the extracted information and finally, based on the parsed questions a particular field is collected and posted as an answer.

\setlength{\tabcolsep}{8pt}  % Increase the separation space between columns.

\begin{table}[b]
\centering
\begin{tabular}{lrr}
\toprule
Method       & \begin{tabular}[c]{@{}c@{}}Answering\\ ANLSL\end{tabular} & \begin{tabular}[c]{@{}c@{}}Retrieval\\ MAP\end{tabular} \\
\midrule
$^{\dagger}$\infrarrd                                   & \textbf{0.7743}   & 74.66\% \\
(Baseline) Database~\cite{doccvqa_icdar}               & 0.7068            & 71.06\% \\
(Baseline) TS-BERT~\cite{doccvqa_icdar}     & 0.4513	        & 72.84\% \\
DQA~\cite{docvqa_das}                       & 0.0000            & \textbf{80.90}\% \\
DOCR~\cite{docvqa_das}                      & 0.0000            & 79.15\% \\
\bottomrule
\end{tabular}
\vspace{1.5mm}
\caption{\textbf{\TaskB task results table.} New submissions after CVPR challenge edition~\cite{docvqa_das} are indicated by $^{\dagger}$ icon.}

\label{tab:task2_results}
\end{table}

\setlength{\tabcolsep}{6pt}  % Restore column separation to its default value.

\subsection{Performance analysis}

\begin{figure}[ht]
\centering
    \includegraphics[width=\linewidth]{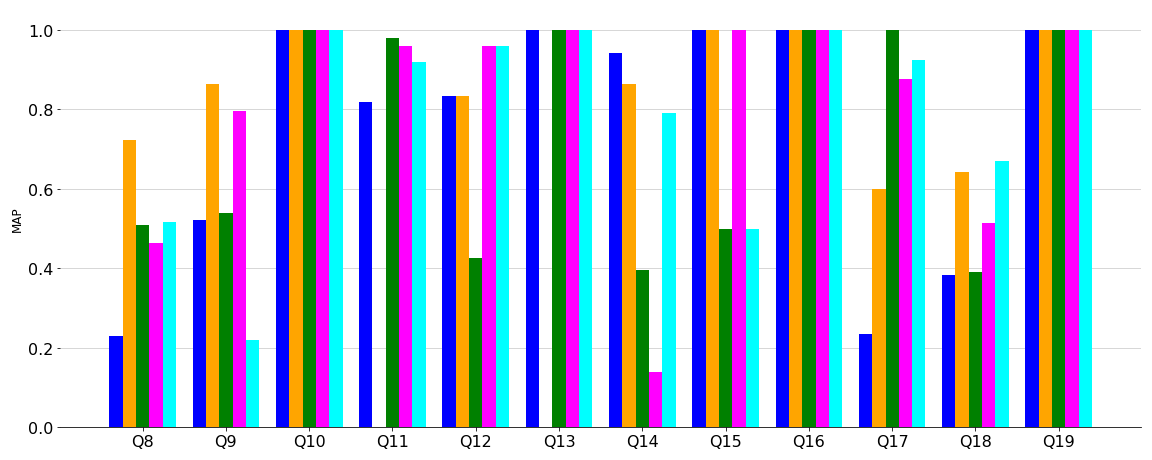}

    \includegraphics[width=\linewidth]{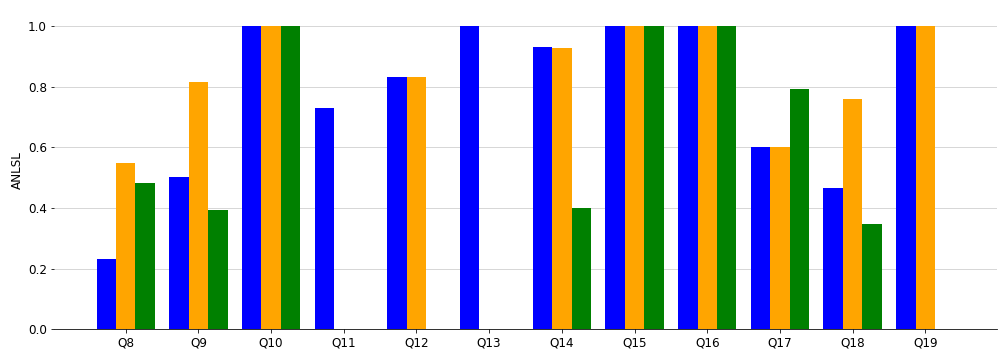}
    
    \includegraphics[width=0.90\linewidth]{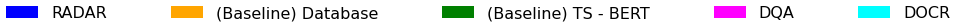}

    \caption{\textbf{\TaskB methods performance by questions}. Top figure shows the retrieval MAP score. Bottom figure shows ANLSL answering score. In bottom figure methods which didn't provide answers are omitted for clarity.}
    \label{fig:t2_scores_by_question}
\end{figure}

Table \ref{tab:task2_results} shows the competition result ranking comparing the submitted method in this challenge with the baselines and the methods from the CVPR 2020 challenge version. The \infrarrd method outperforms all baselines in the ranking metric (ANLSL). However, methods from the 2020 edition show better performance on the retrieval of positive evidence. This indicates a potential for improvement since given the collection nature of this dataset, a better performance in finding the relevant documents is expected to lead to better answering performance. In figure \ref{fig:t2_scores_by_question} we show a breakdown of the evidence and answer scores by query. On one hand there is a set of questions for which most of the methods performs well (Q10, Q15, Q16) while \infrarrd is the only one to come up with the answers for the questions Q11 and Q13 and providing their evidences. However, it performs worse than Database in questions Q8, Q9 and Q18, which is probably a consequence of the performance drop when finding the relevant documents as it can be seen in the top plot. Notice also that some questions refer to only one single document (Q13, Q15, Q16, Q19), which facilitates methods to score either 0 or 1. 

\vspace{-25pt}

\section{\TaskA}
\label{sec:task1}

The \TaskA task was first introduced in the scope of the CVPR 2020 DocVQA Challenge edition. Since then, we have received and evaluated $>750$ submissions from $44$ different users, out of which $28$ have been made public by their authors and feature in the ranking table. $21$ of these submissions are new, and have pushed overall performance up by $\sim2\%$.
In this task questions were posed over single page document images and the answer to the questions was in most of the cases text that appears within the image. The documents of this dataset were sourced from the industry documents library\footnote{\href{https://www.industrydocuments.ucsf.edu/}{https://www.industrydocuments.ucsf.edu/}} and the dataset finally comprised $30,035$ question-answer pairs and $5,485$ images. 

Although the documents feature complex layouts and include tables and figures, running text is dominant. Combined with way questions were defined, extractive approaches inspired from NLP perform quite well. To evaluate the submitted methods the standard ANLS~\cite{biten2019icdar} metric for reading-based VQA tasks was used.

\subsection{Baselines}
\label{subsec:task1_baselines}

In order to establish a baseline performance, we used two models built on state of the art methods presented in~\cite{docvqa_wacv}. 
The first model is the extractive question answering model from NLP BERT~\cite{bert} pretrained on \squad and finetuned on \TaskA dataset. BERT is a language representation pre-trained from unlabeled text passages using transformers. To get the context necessary to extract the answer span we concatenate the detected OCR tokens following left-right, top-down direction. We name this baseline as ``(Baseline) BERT Large'' The second baseline is the same state of the art Scene Text VQA method %model. involving images with embedded scene text. 
\mc~\cite{hu2020iterative} as in \TaskC task. However, since the notion of visual objects in real word images is not directly applicable in case of document images, the features of detected objects are omitted. This baseline is named as ``(Baseline) \mc''.
We also provide a human performance evaluation carried out by few volunteers ``(Baseline) Human''.

\subsection{Submitted methods}
Here we briefly describe the current top 3 ranked methods (new submissions) while the previous edition's methods description can be found in \cite{docvqa_das}. The complete ranking can be found in Table \ref{tab:task1_results}.

\begin{itemize}
\setlength\itemsep{1em}

\item[\#1 -] \tilt~\cite{powalski2021going}: The same team and method as the winners of the \TaskC task.

\item[\#2 -] \layoutlmtwo~\cite{xu2020layoutlmv2}: Pretrained representation that also learns the text, layout and image in a multi-modal framework using new pretraining tasks where cross-modality interaction are better learned than in previous \layoutlm~\cite{xu2020layoutlm}. It also integrates a spatial-aware mechanism that allows the model to better exploit the relative positional relationship among different text blocks.

\item[\#3 -] Alibaba DAMO NLP: An ensemble of 30 models and a pretrained architecture based on StructBERT~\cite{wang2019structbert}.

\end{itemize}

% \vspace{-10pt}

\setlength{\tabcolsep}{8pt}  % Increase the separation space between columns.

\begin{table}[ht]
\centering
\begin{tabular}{lr}
\toprule
Method          & \multicolumn{1}{c}{ANLS} \\
\midrule
(Baseline) Human                & 0.9811 \\
$^{\dagger}$Applica.ai TILT                & \textbf{0.8705} \\
$^{\dagger}$LayoutLM 2.0 (single model)    & 0.8672 \\
$^{\dagger}$Alibaba DAMO NLP               & 0.8506 \\
PingAn-OneConnect-Gammalab-DQA  & 0.8484 \\
Structural LM-v2                & 0.7674 \\
$^{\dagger}$QA\_Base\_MRC\_2               & 0.7415 \\
QA\_Base\_MRC\_1                & 0.7407 \\
HyperDQA\_V4                    & 0.6893 \\
(Baseline) Bert Large           & 0.6650 \\
Bert fulldata fintuned	        & 0.5900 \\
$^{\dagger}$UGLIFT v0.1 (Clova OCR)        & 0.4417 \\
(Baseline) M4C                 & 0.3910 \\
Plain BERT QA                  & 0.3524 \\
HDNet                          & 0.3401 \\
CLOVA OCR                      & 0.3296 \\
DocVQAQV\_V0.1                 & 0.3016 \\
\bottomrule
\end{tabular}
\vspace{2mm}
\caption{\textbf{\TaskA task results table.} New submissions after CVPR challenge edition~\cite{docvqa_das} are indicated by $^{\dagger}$ icon.}
\label{tab:task1_results}
\end{table}

\setlength{\tabcolsep}{6pt}  % Restore column separation to its default value.

% \vspace{-30pt}

\section{Conclusions and Future Work}

The ICDAR 2021 edition of the DocVQA challenge is the continuation of a long-term effort towards Document Visual Question Answering. One year after we first introduced it, the DocVQA Challenge has received significant interest by the community.
The challenge has evolved, by improving evaluation and analysis methods and by introducing a new complex task on \TaskC. In this report, we have summarised these changes, and presented new submissions to the different tasks.
Importantly, we note that while the standard approach one year ago was a pipeline combining commercial OCR with NLP-inspired QA methods, state of the art approaches in 2021 are multi-modal. This confirms that the visual aspect of documents is indeed important, even in more mundane layouts such as the documents of \TaskA. This is especially true when analysing specific types of questions.
The DocVQA challenge will remain open for new submissions in the future. We plan to extend this further with new tasks and insights in the results. We also hope that the VQA paradigm will give rise to more top-down, semantically-driven approaches to document image analysis.

\section*{Acknowledgments}
This work was supported by an AWS Machine Learning Research Award, the CERCA Programme / Generalitat de Catalunya, and UAB PhD scholarship No B18P0070.
We thank especially Dr. R. Manmatha for many useful inputs and discussions.

% This work has been supported by the UAB PIF scholarship B18P0070 and the Consolidated Research Group 2017-SGR-1783 from the Research and University Department of the Catalan Government.
% This work is partly supported by MeitY, Government of India, the project TIN2017-89779-P, 


\bibliographystyle{splncs04}
\footnotesize
\bibliography{main}

\begin{thebibliography}{10}
\providecommand{\url}[1]{\texttt{#1}}
\providecommand{\urlprefix}{URL }
\providecommand{\doi}[1]{https://doi.org/#1}

\bibitem{vqa1}
Agrawal, A., Lu, J., Antol, S., Mitchell, M., Zitnick, C.L., Batra, D., Parikh,
  D.: Vqa: Visual question answering (2016)

\bibitem{topdown_bottomup}
Anderson, P., He, X., Buehler, C., Teney, D., Johnson, M., Gould, S., Zhang,
  L.: {Bottom-Up and Top-Down Attention for Image Captioning and Visual
  Question Answering} (2017)

\bibitem{biten2019icdar}
Biten, A.F., Tito, R., Mafla, A., Gomez, L., Rusinol, M., Mathew, M., Jawahar,
  C., Valveny, E., Karatzas, D.: Icdar 2019 competition on scene text visual
  question answering. In: 2019 International Conference on Document Analysis
  and Recognition (ICDAR). pp. 1563--1570. IEEE (2019)

\bibitem{biten2019scene}
Biten, A.F., Tito, R., Mafla, A., Gomez, L., Rusinol, M., Valveny, E., Jawahar,
  C., Karatzas, D.: Scene text visual question answering. In: Proceedings of
  the IEEE/CVF International Conference on Computer Vision. pp. 4291--4301
  (2019)

\bibitem{leafqa}
{Chaudhry}, R., {Shekhar}, S., {Gupta}, U., {Maneriker}, P., {Bansal}, P.,
  {Joshi}, A.: Leaf-qa: Locate, encode attend for figure question answering.
  In: {WACV} (2020)

\bibitem{bert}
Devlin, J., Chang, M.W., Lee, K., Toutanova, K.: {{BERT}: Pre-training of Deep
  Bidirectional Transformers for Language Understanding}. In: ACL (2019)

\bibitem{drop}
Dua, D., Wang, Y., Dasigi, P., Stanovsky, G., Singh, S., Gardner, M.: {DROP: A
  Reading Comprehension Benchmark Requiring Discrete Reasoning Over
  Paragraphs}. In: NAACL-HLT (2019)

\bibitem{hong2021bros}
Hong, T., Kim, D., Ji, M., Hwang, W., Nam, D., Park, S.: Bros: A pre-trained
  language model for understanding texts in document (2021)  (2021)

\bibitem{hu2020iterative}
Hu, R., Singh, A., Darrell, T., Rohrbach, M.: Iterative answer prediction with
  pointer-augmented multimodal transformers for textvqa. In: Proceedings of the
  IEEE/CVF Conference on Computer Vision and Pattern Recognition (2020)

\bibitem{gqa}
Hudson, D.A., Manning, C.D.: {GQA:} a new dataset for compositional question
  answering over real-world images. CoRR  \textbf{abs/1902.09506} (2019),
  \url{http://arxiv.org/abs/1902.09506}

\bibitem{imagededup}
Jain, T., Lennan, C., John, Z., Tran, D.: {Imagededup}.
  \url{https://github.com/idealo/imagededup} (2019)

\bibitem{triviaqa}
Joshi, M., Choi, E., Weld, D., Zettlemoyer, L.: {{T}rivia{QA}: A Large Scale
  Distantly Supervised Challenge Dataset for Reading Comprehension}. In: ACL
  (2017)

\bibitem{dvqa}
Kafle, K., Price, B., Cohen, S., Kanan, C.: {DVQA: Understanding Data
  Visualizations via Question Answering}. In: CVPR (2018)

\bibitem{figureqa}
Kahou, S.E., Michalski, V., Atkinson, A., K{\'a}d{\'a}r, {\'A}., Trischler, A.,
  Bengio, Y.: Figureqa: An annotated figure dataset for visual reasoning. arXiv
  preprint arXiv:1710.07300  (2017)

\bibitem{textbookqa}
Kembhavi, A., Seo, M., Schwenk, D., Choi, J., Farhadi, A., Hajishirzi, H.: {Are
  You Smarter Than A Sixth Grader? Textbook Question Answering for Multimodal
  Machine Comprehension}. In: CVPR (2017)

\bibitem{natural_quesions}
Kwiatkowski, T., Palomaki, J., Redfield, O., Collins, M., Parikh, A., Alberti,
  C., Epstein, D., Polosukhin, I., Kelcey, M., Devlin, J., Lee, K., Toutanova,
  K.N., Jones, L., Chang, M.W., Dai, A., Uszkoreit, J., Le, Q., Petrov, S.:
  Natural questions: a benchmark for question answering research. Transactions
  of the Association of Computational Linguistics  (2019)

\bibitem{iitcdip_dataset}
Lewis, D., Agam, G., Argamon, S., Frieder, O., Grossman, D., Heard, J.:
  Building a test collection for complex document information processing. In:
  Proceedings of the 29th annual international ACM SIGIR conference on Research
  and development in information retrieval. pp. 665--666 (2006)

\bibitem{visually29k}
Madan, S., Bylinskii, Z., Tancik, M., Recasens, A., Zhong, K., Alsheikh, S.,
  Pfister, H., Oliva, A., Durand, F.: Synthetically trained icon proposals for
  parsing and summarizing infographics. arXiv preprint arXiv:1807.10441  (2018)

\bibitem{infographicvqa}
Mathew, M., Bagal, V., Tito, R.P., Karatzas, D., Valveny, E., Jawahar, C.:
  Infographicvqa. arXiv preprint arXiv:2104.12756  (2021)

\bibitem{docvqa_wacv}
Mathew, M., Karatzas, D., Jawahar, C.V.: {DocVQA: A Dataset for VQA on Document
  Images}. In: WACV (2020)

\bibitem{docvqa_das}
Mathew, M., Tito, R., Karatzas, D., Manmatha, R., Jawahar, C.: Document visual
  question answering challenge 2020. arXiv preprint arXiv:2008.08899  (2020)

\bibitem{ms_marco}
Nguyen, T., et~al.: Ms marco: A human generated machine reading comprehension
  dataset. CoRR  \textbf{abs/1611.09268} (2016)

\bibitem{pasupat2015compositional}
Pasupat, P., Liang, P.: Compositional semantic parsing on semi-structured
  tables. In: Proceedings of the 53rd Annual Meeting of the Association for
  Computational Linguistics and the 7th International Joint Conference on
  Natural Language Processing (Volume 1: Long Papers). pp. 1470--1480 (2015)

\bibitem{powalski2021going}
Powalski, R., Borchmann, {\L}., Jurkiewicz, D., Dwojak, T., Pietruszka, M.,
  Pa{\l}ka, G.: Going full-tilt boogie on document understanding with
  text-image-layout transformer. arXiv preprint arXiv:2102.09550  (2021)

\bibitem{rajpurkar2016squad}
Rajpurkar, P., Zhang, J., Lopyrev, K., Liang, P.: Squad: 100,000+ questions for
  machine comprehension of text. In: Proceedings of the 2016 Conference on
  Empirical Methods in Natural Language Processing. pp. 2383--2392 (2016)

\bibitem{singh2019towards}
Singh, A., Natarajan, V., Shah, M., Jiang, Y., Chen, X., Batra, D., Parikh, D.,
  Rohrbach, M.: Towards vqa models that can read. In: Proceedings of the
  IEEE/CVF CVPR. pp. 8317--8326 (2019)

\bibitem{vqa_tips}
Teney, D., Anderson, P., He, X., van~den Hengel, A.: Tips and tricks for visual
  question answering: Learnings from the 2017 challenge (2017)

\bibitem{doccvqa_icdar}
Tito, R., Karatzas, D., Valveny, E.: Document collection visual question
  answering. arXiv preprint arXiv:2104.14336  (2021)

\bibitem{attention_is_all_you_need}
Vaswani, A., Shazeer, N., Parmar, N., Uszkoreit, J., Jones, L., Gomez, A.N.,
  Kaiser, {\L}., Polosukhin, I.: Attention is all you need. In: Proceedings of
  the 31st International Conference on NeurIPSal Information Processing
  Systems. pp. 6000--6010 (2017)

\bibitem{cocotext}
Veit, A., Matera, T., Neumann, L., Matas, J., Belongie, S.: Coco-text: Dataset
  and benchmark for text detection and recognition in natural images (2016)

\bibitem{wang2019structbert}
Wang, W., Bi, B., Yan, M., Wu, C., Bao, Z., Xia, J., Peng, L., Si, L.:
  Structbert: Incorporating language structures into pre-training for deep
  language understanding. arXiv preprint arXiv:1908.04577  (2019)

\bibitem{xu2020layoutlmv2}
Xu, Y., Xu, Y., Lv, T., Cui, L., Wei, F., Wang, G., Lu, Y., Florencio, D.,
  Zhang, C., Che, W., et~al.: Layoutlmv2: Multi-modal pre-training for
  visually-rich document understanding. arXiv preprint arXiv:2012.14740  (2020)

\bibitem{xu2020layoutlm}
Xu, Y., Li, M., Cui, L., Huang, S., Wei, F., Zhou, M.: Layoutlm: Pre-training
  of text and layout for document image understanding. In: Proceedings of the
  26th ACM SIGKDD International Conference on Knowledge Discovery \& Data
  Mining. pp. 1192--1200 (2020)

\bibitem{recipeqa}
Yagcioglu, S., Erdem, A., Erdem, E., Ikizler-Cinbis, N.: {{R}ecipe{QA}: A
  Challenge Dataset for Multimodal Comprehension of Cooking Recipes}. In: EMNLP
  (2018)

\bibitem{xlnet}
Yang, Z., Dai, Z., Yang, Y., Carbonell, J., Salakhutdinov, R.R., Le, Q.V.:
  {XLNet: Generalized Autoregressive Pretraining for Language Understanding}.
  In: NeurIPS (2019)

\bibitem{zhu2020simple}
Zhu, Q., Gao, C., Wang, P., Wu, Q.: Simple is not easy: A simple strong
  baseline for textvqa and textcaps. arXiv preprint arXiv:2012.05153  (2020)

\end{thebibliography}

\end{document}